\documentclass[conference]{IEEEtran}
\IEEEoverridecommandlockouts
\usepackage[utf8]{inputenc} 
\usepackage[T1]{fontenc}    
\usepackage{hyperref}       
\usepackage{url}            
\usepackage{booktabs}       
\usepackage{amsfonts}       
\usepackage{nicefrac}       
\usepackage{microtype}      
\usepackage{graphicx}			
\usepackage{epsfig}
\usepackage{floatrow}
\usepackage{tabu}
\usepackage{cite}
\usepackage{amsmath,amssymb,amsfonts}
\usepackage{algorithmic}
\usepackage{textcomp}
\usepackage{xcolor}
\usepackage{amsmath}
\usepackage[ruled]{algorithm2e}
\algsetup{linenosize=\small}
\def\BibTeX{{\rm B\kern-.05em{\sc i\kern-.025em b}\kern-.08em
    T\kern-.1667em\lower.7ex\hbox{E}\kern-.125emX}}

\usepackage{graphics} 
\usepackage{times} 
\usepackage{color}
\usepackage{multirow}

\usepackage{xspace}
\makeatletter
\DeclareRobustCommand\onedot{\futurelet\@let@token\@onedot}
\def\@onedot{\ifx\@let@token.\else.\null\fi\xspace}
\let\labelindent\relax

\newcommand{\Cref}[1]{Chap.~\ref{#1}}

\title{Semantic Embedded Deep Neural Network: A Generic Approach to Boost Multi-Label Image Classification Performance}

\author{%
 Xin Shen, Xiaonan Zhao, Rui Luo}
\begin{document}
\maketitle
\begin{abstract}
Fine-grained multi-label classification models have broad applications in e-commerce, such as visual based label predictions ranging from fashion attribute detection to brand recognition. One challenge to achieve satisfactory performance for those classification tasks in real world is the wild visual background signal that contains irrelevant pixels which confuses model to focus onto the region of interest and make prediction upon the specific region. In this paper, we introduce a generic semantic- embedding deep neural network to apply the spatial awareness semantic feature incorporating a channel- wise attention based model to leverage the localization guidance to boost model performance for multi- label prediction. We observed an Avg.relative improvement of 15.27\% in terms of AUC score across all labels compared to the baseline approach. Core experiment and ablation studies involve multi-label fashion attribute classification 
performed on Instagram fashion apparels' image. We compared the model performances among our approach, baseline approach, and 3 alternative approaches to leverage semantic features. Results show favorable performance for our approach. 
\end{abstract}


\section{Introduction}

One big issue for today's industry to develop fashion pattern classification model is to achieve the high precision. The challenges are in 2 aspects: 1) it's a multi-label problem where some patterns are minor to the whole population; 2) the model is working on wild background images (e.g. from Instagram). The multi-label model is used in the following applications: 1) Fashion trend attribute extraction, 2) Image filtering according to fashion attributes, 3) Fashion collection creation. Therefore, its accuracy is critical to the success of the downstream applications. 

Thus, we want to develop a multi-label image classification model that can accurately process images with wild background, considering that some of our applications use images from Instagram. For example, given a fashion image with a fashion model wearing a pinstripe dress standing in front of flowers, we want a network to predict the attribute "pinstripe", but network can be confused by the irrelevant pixels from the wild background of the floral-alike pixels. Therefore, we want the deep learning model to focus on pixels of fashion garment in image, so to have more accurate predictions. To achieve this, we need the classification model to be guided by regional semantic information.
In one previous work of from Mahdi M. Kalayeh et al.\cite{Scheurer2020SemanticSO}, authors concatenate semantic segmentation mask with raw image as input feature, so the whole CNN layers are aware of regional semantic information, which significantly improves classification performance.
However, this approach requires a semantic segmentation model with pixel level annotation, which cannot be used when such labels are not available. 

Following the idea, we propose to solve this problem with an innovative model that has 3 components: 1) We use classification label to train a image classifier, but use it as class activation map (CAM) generator, which contains regional information. 2) We use CAM as an input to learn a semantic embedding for each pixel, where regional semantic information is encoded. 3) We concatenate semantic embedding with the raw image tensor as input features and feed them into a channel wise attention based image classification model to enhance the fine-grained multi-label prediction. 

\begin{figure*}[t]
    \centering
    \includegraphics[width=0.8\textwidth]{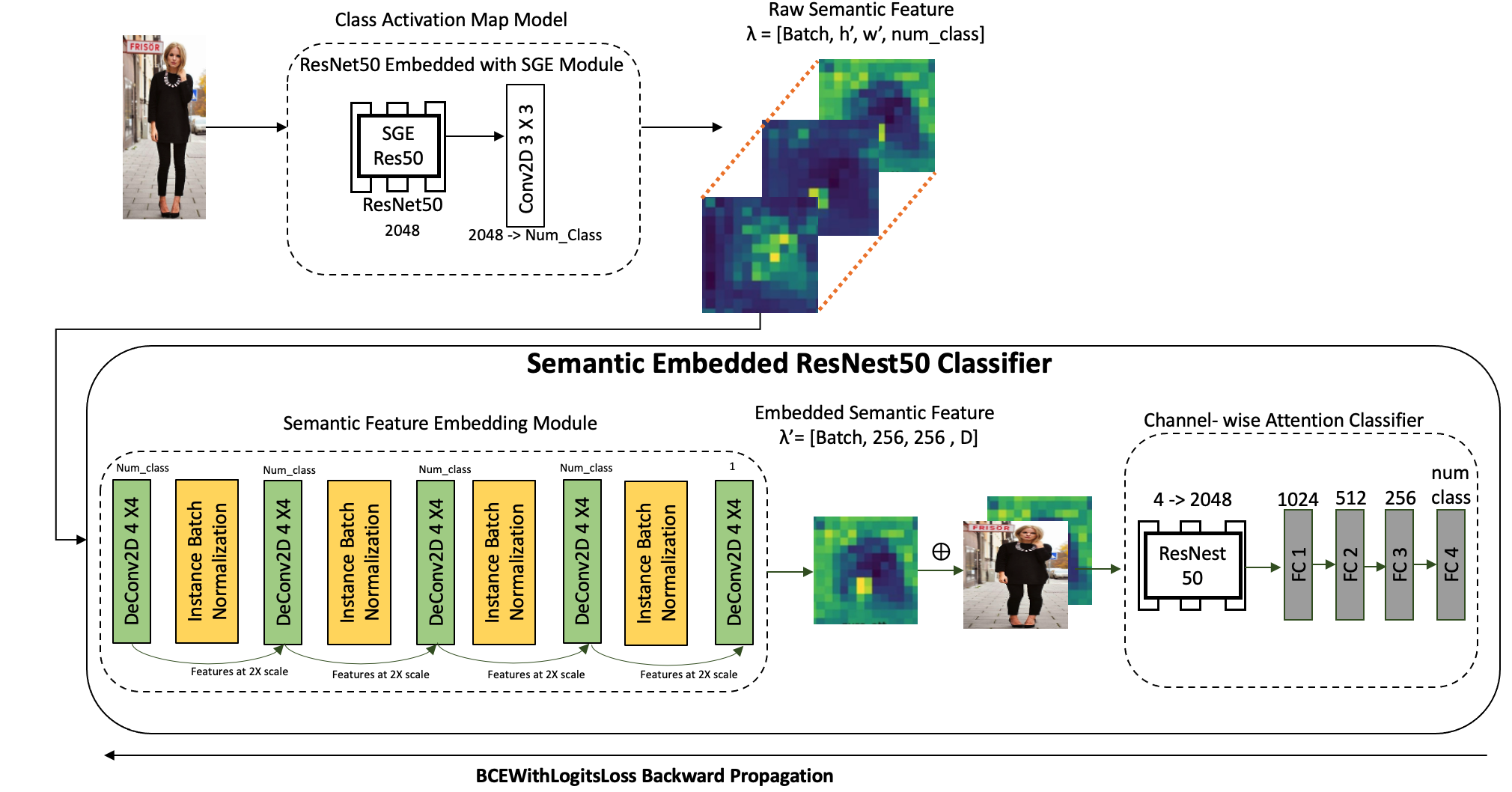}
    \caption{The model structure of the proposed Semantic Embedded ResNest50 Classifier}
    \label{fig:embedded_resnest}
\end{figure*}

\section{Related Work}

\paragraph{Semantic guidance} Semantic guidance,i.e. using semantic information as an extra input or feature to guide feature extraction or classification, are widely adopted to boost target tasks, such as classification, visual search, GAN based image rectification and semantic based regression~\cite{shen2023fishrecgan, shen2023learning}, crowd counting\cite{Xu2019CrowdCW}, etc. Some previous work\cite{Scheurer2020SemanticSO, Takahama2019MultiStagePI} uses semantic information as input of CNN networks to train a CNN classifier. 
Some previous works perform attentions between semantic segmentation output and feature maps from CNN backbone \cite{FIDALGO2018BoostingIC, Li2018TellMW, Li2020SelfSupervisedVA, Kalayeh2018HumanSP}, but these methods rely on a supervised semantic segmentation model to generate the semantic segmentation map. 
There are some research works using class activation maps to boost the CNN model to extract fine-grained visual features\cite{Yang2019TowardsRF}.
Human semantic parsing\cite{Kalayeh2018HumanSP} is used for person-reID. Dense pose\cite{Neverova2018DensePT} are used to train generative models for outfit transfer or outfit generation\cite{Lassner2017AGM}. In the work of crow counting\cite{Xu2019CrowdCW}, semantic segmentation map is used to correct the final prediction from the regression model.

\paragraph{Multi-label classification}
Multi-label classification problem is hard mainly because of the nature of imbalanced distribution. Some previous works also leveraged semantic information to help the problem. In the work\cite{Zhang2018MultilabelIC}, regional latent semantic dependencies are learned by a mixer of CNN and LSTM model. Similarly, in their work\cite{Xu2021JointIA}, they uses Mask R-CNN model to generate object bounding boxes, then perform a graph based multi-label classifier. To our best knowledge, all these works rely on a pre-trained semantic segmentation model or an object detection model, which has same semantic dictionary\cite{Zhou2021DeepSD,Jing2016MultiLabelDL} as the target task. This is a major drawback for generalization on any tasks.

\section{Method}
\begin{figure}[h]
    \centering
    \includegraphics[width=0.75\textwidth]{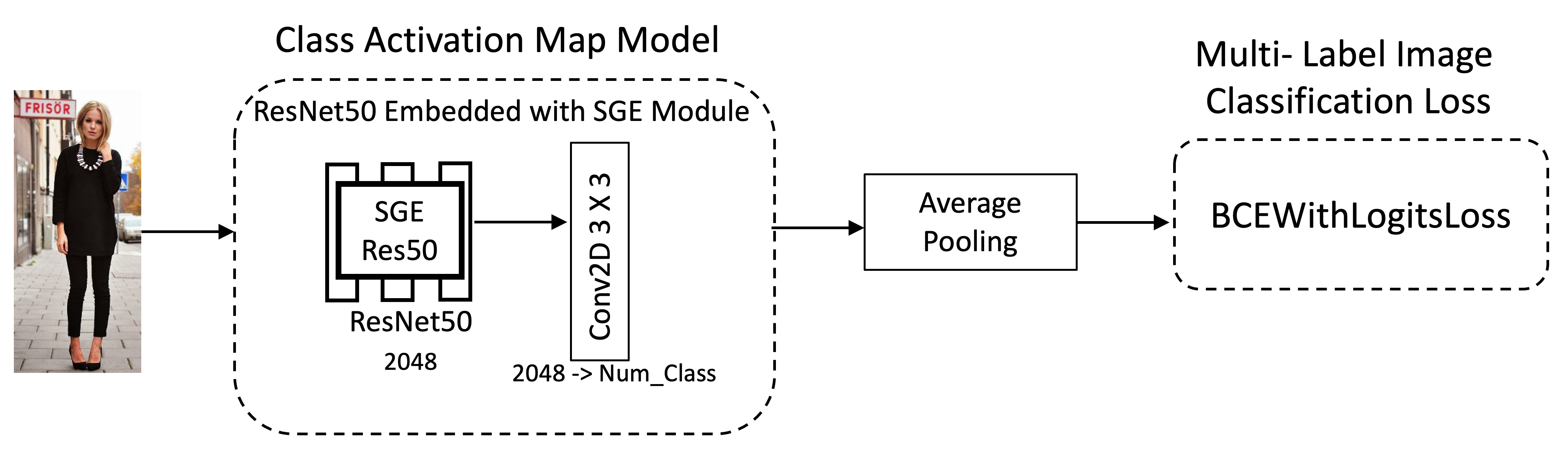}
    \caption{Model structure and training scheme of the semantic generator}
    \label{fig:semantic_generator}
\end{figure}

Firstly, we fine-tune a pre-trained ResNet50\cite{he2016deep} incorporated with attention-based SGE module\cite{li2019spatial} with multi-label Sigmoid Cross-Entropy loss (denoted as BCEWithLogitsLoss in PyTorch). We added a convolution layer after Resnet50 to transform the feature map from $\mathbb[h', w', 2048]$ to $\mathbb[h', w', c]$, where c equals to number of classes (see Fig. \ref{fig:semantic_generator}). The class activation map ($\mathbb[h', w', c]$) is used as input to generate semantic embeddings in the next component.

As shown in Fig.\ref{fig:embedded_resnest}, we utilize CAM Model to generate a class activation map (CAM) $\mathbb{\lambda}$ in the size of $\mathbb[Batch, W, H, {num}\_{class}]$. We then feed the CAM into a Semantic Feature Embedding Module - a 5-layer of deconvolution network. Instance normalization \cite{ulyanov2016instance} is plugged in deconvolution layers to normalize the output of regional semantic embedding ${\lambda}^{'}$. The regional semantic embedding ${\lambda}^{'}$ from the embedding layer in the size of $\mathbb [Batch, 256, 256, D]$ is concatenated with the raw image as extra $D$ channel(s). 

The concatenated tensor is fed into the channel-wise attention based multi-label image classifier (ResNest50\cite{zhang2020resnest}). We found that channel wise attention is better than plain convolutions, when it learns the correlation between the regional semantic embedding and the raw RGB channels. We further added more fully connected layers (FCs) to introduce more non- linearity.

\section{Experiments}
\subsection{Dataset Description}
\begin{figure}[h]
    \centering
    \includegraphics[width=0.6\textwidth]{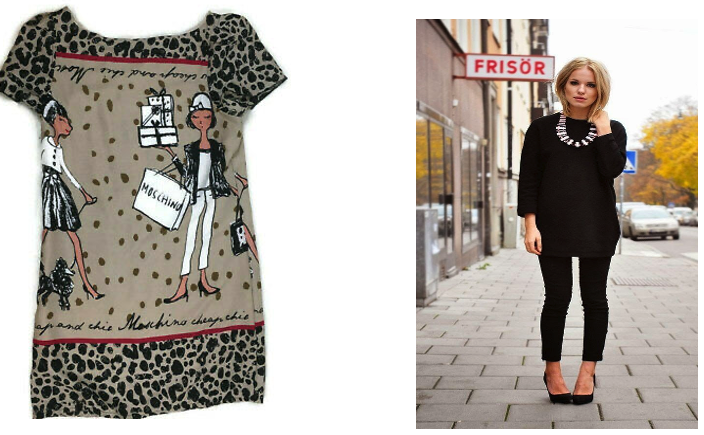}
    \caption{The left image shows a women's shirt containing pattern attributes of "Animal" and "Dots" and "Graphics". The right image shows a sample of the image domain we develop and test the model on, where the background behind the model is quite noisier, which makes it more challenging for a DNN model to predict the pattern attribute from such images.}
    \label{fig:image_sample}
\end{figure}

\begin{figure}[h]
    \centering
    \includegraphics[width=0.75\textwidth]{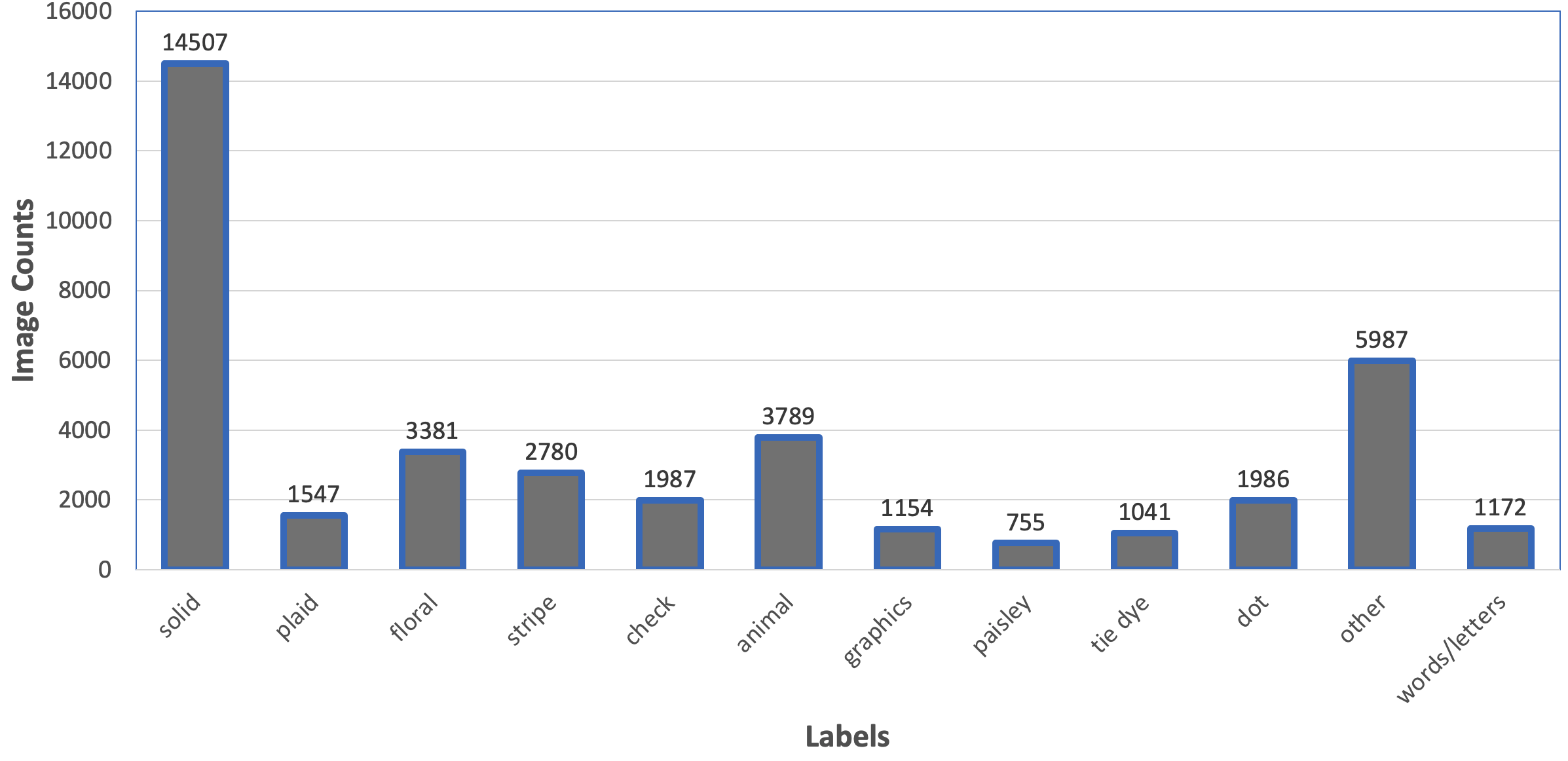}
    \caption{The label distribution visualization}
    \label{fig:label_distribution}
\end{figure}

We collected 34,158 Instagram fashion apparel images for women's dressing, such as dress, tops, and bottoms where the image contains a wilder background noise, shown in the right figure of Fig.\ref{fig:image_sample}. Images are annotated in a multi-label fashion with 11 pattern attributes: \textit{\{Solid, Plaid, Floral, Stripe, Check, Graphic, Tie Dye, Animal, Words/Letters, Dot, Paisley\}}.The dataset shows a severe label imbalance issue, see Fig.\ref{fig:label_distribution}. These images were randomly partitioned into training(70\%) and testing(30\%).


\subsection{Problem Setting and Evaluation Metrics}
\begin{figure}[h]
    \centering
    \includegraphics[width=0.95\textwidth]{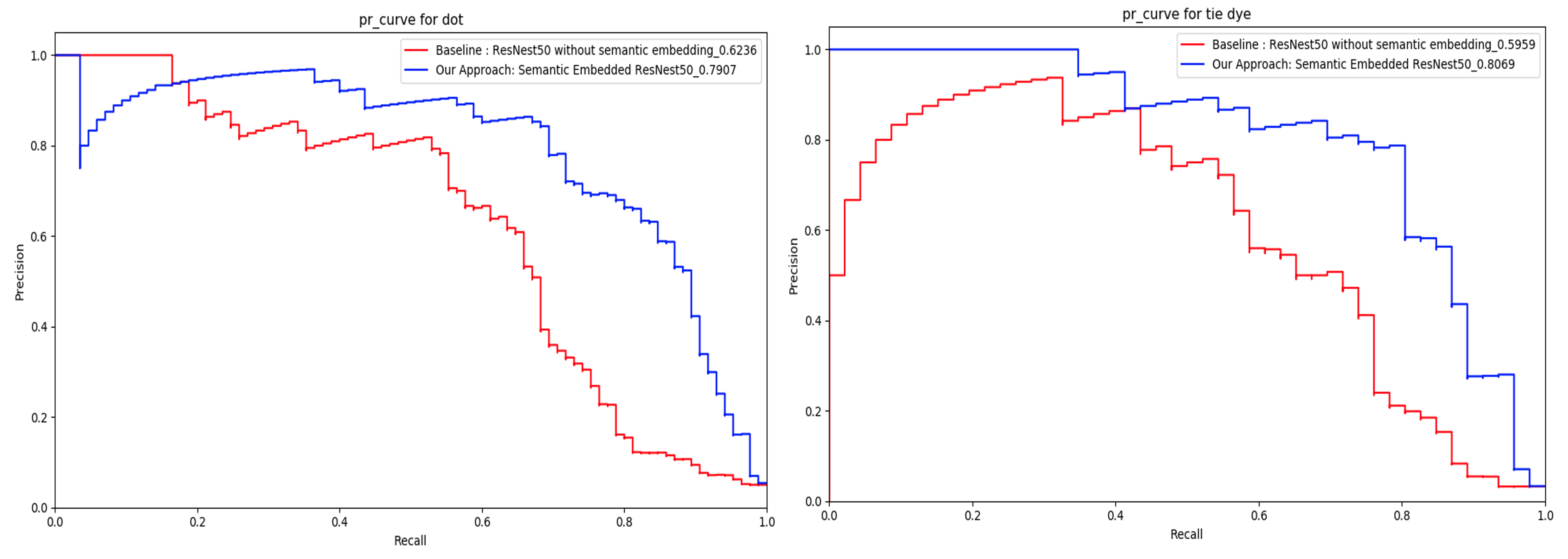}
    \caption{Sample of Model Performance Visualization of PR- Curve. The number in the legend manifests this approach's AUC score}
    \label{fig:PR}
\end{figure}

We defined this visual detection work as a multi-label classification problem, given the nature of the fashion attributes which could co-exist in a same women's dressing, as shown in the left figure of Fig.\ref{fig:image_sample}. In turn the ground truth is labeled into the form of multi-label one-hot vector, such as \textit{[1,0,0,0,0,0,1,1,0,0,0]}, and the output from the model is the corresponding probability distribution for each label in the same dimension as the ground truth. The model performance is evaluated using the area under the curve (AUC) of precision-recall curve on the test set for each label shown in Fig.\ref{fig:PR}



\subsection{Network and Training}
We selected 2 models for experimental comparisons: 1) the baseline model is a ResNest50 \cite{zhang2020resnest} without applying any semantic guidance (referred as \textit{ResNest50}), 2) our Semantic Embedded ResNest50 approach (referred as \textit{Semantic Embedded ResNest50}). Model parameters can be found in Fig.\ref{fig:embedded_resnest}. We also found that increasing the output channel size from 1 to 3 of the last deconvolution layer does not consistently improve the model performance (see Ablation Study). Thus, we set the output channel size to be 1. We trained models using Sigmoid Cross-Entropy loss. We reduced learning rate when the loss has stopped improving after 4 epochs. The learning rate was initiated to be $1e^{-5}$ and Adam Optimizer\cite{kingma2014adam} was adopted.

\section{Results and Discussion}
Table.\ref{table:archPeromance} compares the model performance between baseline approach (plain ResNest50) and our approach (Semantic Embedded ResNest50). Our approach out-performed the baseline approach across ALL labels in terms of the AUC score, with relative improvement ranging from  \textbf{7.54\%} to \textbf{35.41\%}. This manifests using semantic embedding as a guidance significantly boosts up the model performance as the extra localization information embedded helps a network to focus more onto the region of interest and ignore irrelevant pixels. 

\begingroup
\setlength{\tabcolsep}{0.1pt} 
\renewcommand{\arraystretch}{3} 
\begin{table}[H]
    \centering
    \fontsize{5}{8}\selectfont \setlength{\tabcolsep}{0.3em}
    \caption{Comparison on the label-wise AUC performance: 1). Baseline: ResNest50 with raw RGB input, 2).Our approach: Semantic Embedded ResNet50 leveraging semantic embedding}    
    \label{table:archPeromance}
    \begin{tabular}{|| c | c | c | c| c | c | c | c | c | c | c | c ||}
    \hline
       & \textbf{Solid} & \textbf{Plaid} & \textbf{Floral} & \textbf{Stripe}  
       &\textbf{Check} &\textbf{Animal} & \textbf{Graphics} & \textbf{Paisley} 
       &\textbf{\shortstack{Tie \\ Dye}} &\textbf{Dot} &\textbf{\shortstack{Words \\Letters}}\\
        \hline
          \textbf{~~~ResNest50~~~}  &  0.809 & 0.747 & 0.621 & 0.704 & 0.732 & 0.786 & 0.468 & 0.440 & 0.596 & 0.624 & 0.555 \\
        \hline
        
        \hline
        \textbf{\shortstack{Semantic \\ Embedded \\ ResNest50}} & {\color{black}0.870} & {\color{black}0.840} & {\color{black}0.740} & {\color{black}0.793} & {\color{black}0.791} & {\color{black}0.880} &
        {\color{black}0.509} & {\color{black}0.549} & {\color{black}0.807} & {\color{black}0.791} & {\color{black}0.626}\\
        \hline
        
        \hline
        \textbf{\shortstack{Relative \\ Improvement}} & {\color{blue}7.54\%} & {\color{blue}12.40\%} & {\color{blue}19.19\%} & {\color{blue}12.59\%} & {\color{blue}7.99\%} & {\color{blue}11.98\%} &
        {\color{blue}8.83\%} & {\color{blue}24.84\%} & {\color{blue}35.41\%} & {\color{blue}26.80\%} & {\color{blue}12.75\%}\\
        \hline
    \end{tabular}
\end{table}
\endgroup

We consider the majority class to be the label with image samples over 5\% and the rest are minorities. In turn, the majority classes are \textit{Solid}, \textit{Stripe} and \textit{Animal}, while minorities are \textit{Plaid, Check, Graphics, Paisley, Tie Dye, Dot, Words/Letters, and Floral}. We compared the average relative improvement across all majorities and all minorities respectively. According to Table.\ref{table:majority_minority}, our approach yields a higher improvement on minority classes (image samples under 5\%) than it performs on majorities. From this perspective, our approach can be leveraged to alleviate label imbalanced issue. 

\begingroup
\setlength{\tabcolsep}{1.8pt} 
\renewcommand{\arraystretch}{3} 
\begin{table}[H]
    \centering
        \caption{Avg. Relative Improvement over majority classes and over minority classes}  
    \begin{tabular}{|| c | c ||}
    \hline
    \textbf{\shortstack{Avg. Relative Improvement \\ over Majorities}} & {11.45\%} \\
    \hline
    \textbf{\shortstack{Avg. Relative Improvement \\ over Minorities}}&  {\color{blue}16.64\%}\\
    \hline
    \end{tabular}
  
    \label{table:majority_minority}
\end{table}
\endgroup

\begin{figure*}[t]
    \centering
    \includegraphics[width=0.75\textwidth]{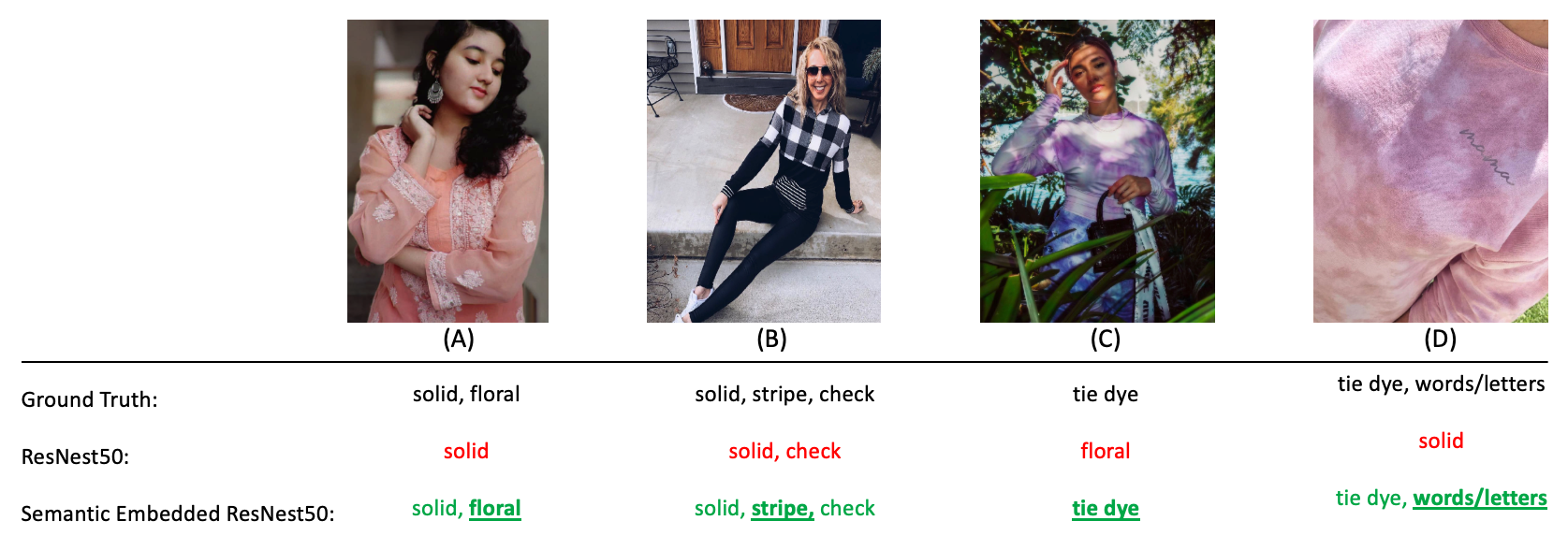}
    \caption{The visualization showing the comparison between the baseline's approach and our approach. All four images contain at least 1 minority class and might/ might not contain majority class. For example, Image.A contains one majority class as solid, and one minority class as floral}
    \label{fig:prediction_comparison}
\end{figure*}

Further, Fig.\ref{fig:prediction_comparison} visualizes four image samples with the multi-label predictions made by the baseline approach and our approach respectively. From the prediction results, our approach shows a preciser prediction than the baseline approach and it could better capture the pattern of minority classes. In turn when an image contains both majority class and minority class, our approach does not solely focus onto majority class but also makes a correct prediction for minority class (see Image.A and Image.B). Meanwhile, we observed that for label "tie dye", the baseline approach tends to be confused by the wild background pixels (see Image.C) and by the less-strong and insignificant pixel signal of "tie dye" (see Image.D), while our approach could successfully extract the needed pixels for a correct prediction.

\subsection{Ablation Study}
The classification module is composed with 2 modules: 1) Semantic Embedding Module which serves to embed a raw CAM into a semantic embedding with the shape of $\mathbb[Batch, 256, 256, 1]$, and 2). ResNest50 as the classification backbone to provide the network with a channel- wise attention mechanism. We conducted 3 following ablation studies: 1) we compared our Semantic Embedded Module to embed a raw CAM with a simple interpolation based tensor-reshape. We performed tensor reshape and a channel-wise summation with min-max normalization to achieve a 1-D feature map in the shape of $\mathbb[Batch, 256, 256, 1]$, 2) we wanted to understand if a plain convolution network without channel- wise attention, Resnet50, can be used to substitute the ResNest50 (referred as Semantic Embedded Resnet50). 3) we tested if a 3- channel embedded semantic feature generated by Semantic Embedding Module could achieve a better model performance (referred as 3-channel Semantic Embedded ResNest50) compared to 1-D semantic embedding.  

\begingroup
\setlength{\tabcolsep}{0.3pt} 
\renewcommand{\arraystretch}{3} 
\begin{table}[H]
    \centering
    \fontsize{5}{8}\selectfont \setlength{\tabcolsep}{0.3em}
    \caption{Comparison on the label-wise AUC performance between our approach Semantic Embedded ResNest50 and all other 3 ablation studies  models: 1) Manually Reshaped Semantic, 2) Semantic Embedded Resnet50, 3) 3-channel Semantic Embedded ResNest50}    
    \label{table:ablation}
    \begin{tabular}{|| c | c | c | c| c | c | c | c | c | c | c | c ||}
    \hline
      & \textbf{Solid} & \textbf{Plaid} & \textbf{Floral} & \textbf{Stripe}  
      &\textbf{Check} &\textbf{Animal} & \textbf{Graphics} & \textbf{Paisley} 
      &\textbf{Tie Dye} &\textbf{Dot} &\textbf{\shortstack{Words/ \\Letters}}\\
        \hline
        \textbf{\shortstack{ResNest50 \\ Deconv D=1}} & {\color{black}0.870} & {\color{black}0.840} & {\color{black}0.740} & {\color{black}0.793} & {\color{black}0.791} & {\color{black}0.880} &
        {\color{black}0.509} & {\color{black}0.549} & {\color{black}0.807} & {\color{black}0.791} & {\color{black}0.626}\\
        \hline
        
        \textbf{\shortstack{ResNest50 \\ Reshaped D=1}} & {\color{red}-0.387} & {\color{red}-0.726} & {\color{red}-0.620} & {\color{red}-0.670} & {\color{red}-0.350} & {\color{red}-0.768} &
        {\color{red}-0.478} & {\color{red}-0.523} & {\color{red}-0.785} & {\color{red}-0.740} & {\color{red}-0.601}\\
        \hline
        
        \textbf{\shortstack{Resnet50 \\ Deconv D=1}}& {\color{red}-0.028} & {\color{red}-0.0366} & {\color{red}-0.024} & {\color{red}-0.013} & {\color{red}-0.018} & {\color{red}-0.013} &
        {\color{red}-0.021} & {\color{red}-0.027} & {\color{red}-0.025} & {\color{red}-0.024} & {\color{red}-0.1072}\\
        \hline
        
        \textbf{\shortstack{ResNest50 \\ Deconv D=3}}& 
        {\color{red}-0.093} & {\color{blue}+0.008} & {\color{red}-0.006} & {\color{red}-0.040} & {\color{blue}+0.011} & {\color{blue}+0.019} &
        {\color{red}-0.145} & {\color{red}-0.056} & {\color{blue}+0.014} & {\color{blue}+0.016} & {\color{red}-0.032}\\
        \hline
    \end{tabular}
\end{table}
\endgroup

The comparison between $2^{nd}$ row and $1^{st}$ row from Table.\ref{table:ablation} shows a heavily decayed performance for the approach of concatenating a manually reshaped and normalized semantic map. This manifests this approach harms the overall learning progress. The interpolation based reshaping mechanism and channel- wise summation to reshape a raw CAM from a dimension of \textit{[w', h', c]} into \textit{[256, 256, 1]} introduces many noises to the original CAM. Still this manually work is not learnable, thus, at the inference stage, the extra information provided by this manually reshaped semantic feature is in a random pattern to a DNN. This introduces more irrelevant noises for a DNN to process and in turn hurts the model performance. 

Further, the comparison between Semantic Embedded Resnet50 and Semantic Embedded ResNest50, shown in $3^{rd}$ row and $1^{st}$ row, manifests this alternative approach fails to yield a boosted performance. The average relative decrease in terms of AUC score across all labels using this approach compared to our approach's is \textbf{4.39\%}. These manifest that our approach, Semantic Embedded ResNest50 which leverages the channel- wise attention mechanism, is more sensitive to the semantic embedding guidance. The channel- wise attention mechanism enables a soft-attention-like effect via the semantic embedding guidance to help the classifier to learn the localization information better and in turn help the classifier to focus more onto the region of interest for a better model performance.

Lastly, the last row of Table.\ref{table:ablation} shows the model performance comparison between our approach of using a 1-channel semantic embedding and an alternative approach of using a 3-channel semantic embedding. The mixed performance manifests that there is no significant correlation between the channel size of the semantic embedding and model performance. The Semantic Embedded Module we built, shown in Fig.\ref{fig:embedded_resnest}, expands the input raw CAM's dimension by 2X without changing the channel size over the first 4 deconvolution layers, and the last deconvolution layer performs both dimension reshaping and channel size squeezing. In turn, the localization information from raw CAM is both retained throughout the entire 5 deconvolution layers no matter whether the last layer convolves the feature into a 3 channel or 1 channel.That is, the 1-channel semantic embedding embeds the information more densely, while the semantic information from a 3-channel semantic embedding is more diluted and distributed across 3 channels. 

\section{Industrial Impact}

Our work presents a generic deep neural network structure to build a multi-label classification model with a high precision and recall. This model serves as a crucial upper stream for many down stream applications across E-commerce, such as Fashion Industry. For example, to understand how a fashion attribute evolve over a time period, we need the temporal signal of a certain attribute over a time period, and our work can contribute to make attribute signal extraction and aggregation. Further, this model can be used as image content understanding for customer search query to products matching on the fashion attribute level. For example, when a customer search a query of "Floral Tie Dye Shirt", most existed query to product matcher mechanism is based on the similarity between query and  product-titles. However, a free text based signal might contain more noises and even fake attribute signal (sellers tend to include more attribute information to make a title attractive) compared to pure visual signal representation. Hence, our model can serve as a proficient encoder for semantic embeddings, providing valuable support to a wide range of downstream applications. Examples of such applications include text embedding/ image embedding based query-to-product matching and sequential-based item retrieval, where our model's capabilities can significantly enhance the overall performance and effectiveness.~\cite{chen2021efficient, wsfe, li2023text}. More importantly, we introduced a generic deep learning approach to achieve a high- performance backbone more than fashion detection domain. Many other industries, such as mechanical industry, could leverage our approach to build their own high- performance backbone to operate tasks including but not limited to object's relocation and deviation detection. By reframing the conventional approach of object status detection using traditional geometric based computer vision into a machine learning-based classification and regression problem, our solution offers a compelling alternative to existing works~\cite{mei2017mechanics, mei2018erratum} that heavily depend on costly hardware. In doing so, we introduce an efficient methodology that addresses this challenge with remarkable efficacy. 
\section{Conclusion}
Overall, by leveraging our proposed approach, Semantic Embedded ResNest50, enables us to obtain a boosted model performance in terms of fine grained multi- label prediction. The trainable semantic embedding layer enables a dynamic mechanism to reshape and to embed a raw CAM into the desired shape without introducing extra noises and losing signals during the process of deconvolution, while the channel- wise attention based classifier backbone enables a soft- attention- like mechanism to leverage the localization information provided by the embedded semantic map to help the network to focus more onto the region of interest from the RGB channel. Experimental results on fashion pattern attribute classification show significant boost using this proposed approach with a well- rounded improvement across all labels ranging from 7.54\% to 35.41\% in terms relative AUC improvement compared to the baseline approach. Moreover, our approach shows a favorable performance to the minority classes, which improves a model's robustness upon the minority classes. To further improve the model's generalization abilities upon minority classes, we could incorporate label distribution stabilization techniques~\cite{shen2023data, liu2021two}. 

Our next step is to investigate an end to end solution from generating the raw CAM to embed the semantic embedding, and to the prediction layer using one single DNN model. This merged deep learning approach could remove the training efforts of a separate CAM generator, which in turn removes a dependency. 

\bibliographystyle{plain}
\bibliography{references}


\end{document}